\documentclass{bmvc2k}
\usepackage{svg}
\usepackage{booktabs}
\usepackage{adjustbox}
\usepackage{amssymb}  
\usepackage{bmvc2k_natbib}

\title{GHR-VQA: Graph-guided Hierarchical Relational Reasoning for Video Question Answering}

\addauthor{Dionysia Danai Brilli}{}{1}
\addauthor{Dimitrios Mallis}{}{3}
\addauthor{Vassilis Pitsikalis}{}{4}
\addauthor{Petros Maragos}{}{1,2}

\addinstitution{
 School of ECE\\
 National Technical University of Athens\\
 Athens, Greece
}
\addinstitution{
Robotics Institute\\
Athena Research Center\\
Athens, Greece
}
\addinstitution{
University of Luxembourg, \\
Kirchberg, Luxembourg}
 
\addinstitution{
deeplab.ai,\\
Athens, Greece
}

\runninghead{BRILLI, MALLIS, PITSIKALIS, MARAGOS}{GHR-VQA}

\begin{document}

\maketitle

\begin{abstract}
We propose GHR-VQA, Graph-guided Hierarchical Relational Reasoning for Video Question Answering (Video QA), a novel human-centric framework that incorporates scene graphs to capture intricate human-object interactions within video sequences. Unlike traditional pixel-based methods, each frame is represented as a scene graph and human nodes across frames are linked to a global root, forming the video-level graph and enabling cross-frame reasoning centered on human actors. The video-level graphs are then processed by Graph Neural Networks (GNNs), transforming them into rich, context-aware embeddings for efficient processing. Finally, these embeddings are integrated with question features in a hierarchical network operating across different abstraction levels, enhancing both local and global understanding of video content. This explicit human-rooted structure enhances interpretability by decomposing actions into human-object interactions and enables a more profound understanding of spatiotemporal dynamics.  We validate our approach on the Action Genome Question Answering (AGQA) dataset, achieving significant performance improvements, including a 7.3\% improvement in object-relation reasoning over the state of the art.
\end{abstract}

\section{Introduction}
\label{sec:intro}
In the rapidly evolving digital era, the exponential growth of video content presents unprecedented opportunities for understanding human behavior. From everyday activities to complex interactions, videos capture rich sequences of humans interacting with their environment and others. Extracting meaningful knowledge from these sequences requires more than recognizing objects or isolated actions; it needs reasoning about the relationships and dynamics that unfold over time. Video Question Answering (VideoQA) emerges as a powerful framework for this challenge, enabling machines to understand and reason about temporal and spatial dynamics within videos to answer natural language questions.

\begin{figure*}[t]
  \centering
   \includegraphics[width=0.95\linewidth]{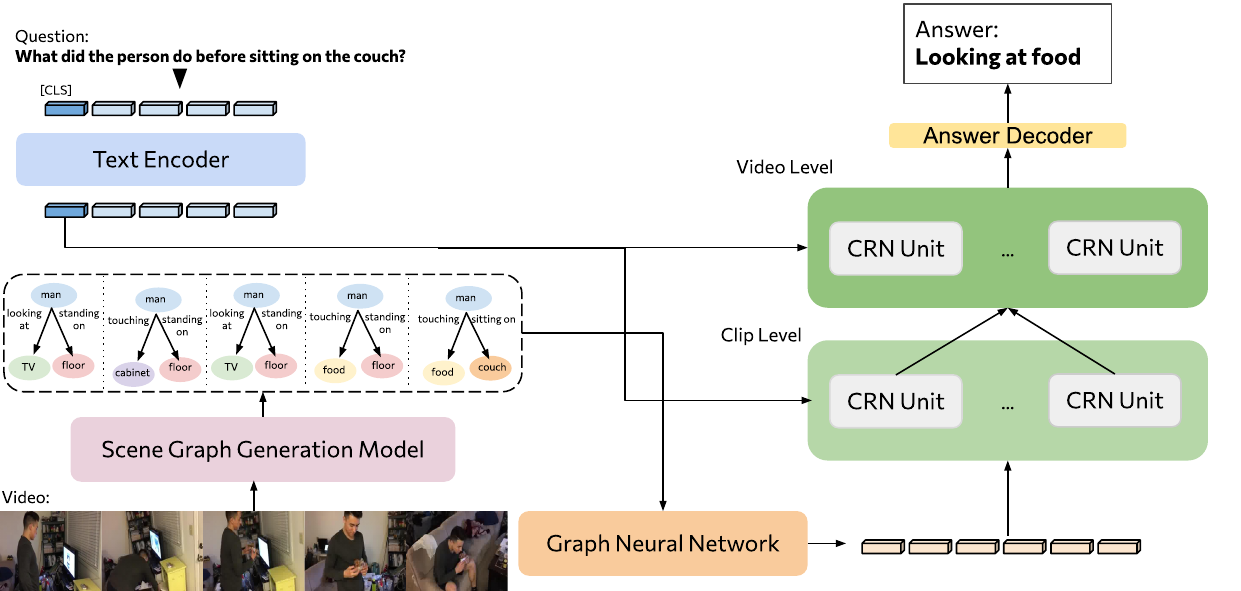}
   \caption{ Our proposed architecture. The process begins with the input of a question and a corresponding video. Initially, we perform clip selection and pass the segments through an SGG model to extract scene graphs that represent the visual elements and their interrelationships. These extracted scene graphs are processed by a GNN, which generates meaningful embeddings. The embeddings are then fed into a hierarchical network, which integrates and contextualizes the information across different levels of abstraction to generate a comprehensive understanding in relation to the query and finally answer the question.}
   \label{fig:onecol}
\end{figure*}

Despite continuous advancements, the field still faces significant challenges. A key difficulty lies in effectively capturing the intricate interconnections between entities in a scene and their evolution over time. Many approaches treat humans, objects and their relationships as secondary and implicit features within a global video representation. This means that human-object interactions are not explicitly represented as structured entities, but are embedded implicitly in video feature maps. As a result, models may learn to answer questions by exploiting dataset biases or global context rather than reasoning over the specific human-object interactions within a scene. Also, the temporal dimension increases complexity, with existing approaches often struggling to capture and interpret dynamic object relationships within videos \cite{Paul2023}. A line of research ~\cite{khan_learning_2023} explores learning situation hyper-graphs for videos by extracting intermediate latent graph representations from raw visual scenes. While these methods enable effective modeling of entities and their relationships, they are highly resource-intensive as they involve complex architectures and require extensive annotated datasets, which may not always be readily available.

We address this gap by introducing a novel modeling approach for VideoQA, in which each video is represented as a human-rooted video-level graph. Each frame is first expressed as a scene graph and human nodes across frames are linked to a global root node. This structure allows cross-frame reasoning centered on humans and enables targeted and efficient message passing on explicit human-object relationships. By leveraging an off-the-shelf Scene Graph Generation (SGG) model, we infer scene graphs from video frames, eliminating dependence on raw video frames. Instead, as highly compact  and interpretable representations, scene graphs facilitate lightweight yet effective modeling of spatiotemporal dynamics.

Our pipeline, GHR-VQA, begins with an efficient Scene Graph Generation (SGG) step using the state-of-the-art method from ~\cite{motifs}. Instead of treating each frame independently, we convert the sequence of scene-graphs into a single human-rooted video-level graph by connecting all frame-level graphs through a shared global root node. This root node serves as a central hub for cross-frame communication, allowing information to travel across objects and scenes. The generated graphs serve as input to the Scene Graph Encoding Module (SGEM), which maps them to latent representations through a shallow Graph Neural Network (GNN). These graph representations, combined with BERT-based question encodings, are then processed by a hierarchical network to produce conditioned embeddings and the final answer. We evaluated our method in the Action Genome Question Answering Dataset \cite{agqa}, a challenging benchmark designed for compositional spatiotemporal reasoning, where we demonstrate robust performance.

\section{Related Works}
\subsection{Video Question Answering}

Video Question Answering is an emerging task at the intersection of computer vision and natural language processing. It requires models to reason about spatiotemporal information in videos to answer questions correctly. Early approaches relied on recurrent neural networks (RNNs) and convolutional neural networks (CNNs) to process temporal and spatial features, respectively \cite{rnn, cnn}. More recent advances have incorporated attention mechanisms and transformers, enabling long-range dependency modeling and improved multimodal reasoning \cite{zhao2017video, Fan_2019_CVPR}. Many contemporary techniques approach this task with Large Language Models (LLMs) and Large Visual-Language Models (LVLMs) \cite{Pan_2023_ICCV, kim2024imagegridworthvideo}. These pre-trained models learn from a massive amount of diverse data and can handle unseen data effectively. Despite these advancements, existing methods often struggle with capturing detailed object interactions and higher-level reasoning, and often rely upon language biases, motivating research into structured representations\cite{ji2020action}.

\subsection{Graphs in Video QA}
A promising direction of Video Question Answering is using graph-based methods, where relationships between objects, actions and attributes are explicitly modeled. Graph-based models offer advantages in interpretability and relational reasoning but can be computationally expensive and reliant on accurate object detection. Many methods leverage scene-graphs, which represent objects, relationships and attributes as structured triplets, since they capture semantic relationships explicitly \cite{dang2021hierarchical, hu2019language, xiao2022video}. Mao et al. introduced dynamic multistep reasoning over video scene graphs, iteratively aligning question tokens with graph nodes and edges to perform multi-hop inference across time ~\cite{mao2022dynamic}. Liu et al. (HAIR) constructed hierarchical visual–semantic graphs with a graph memory to aggregate frame-level object nodes into higher-level representations for VideoQA ~\cite{liu2021hair}. Bai et al. proposed event graph guidance for compositional spatio-temporal reasoning, structuring video into event nodes and relations to answer multi-step queries~\cite{bai2024event}. There have also been attempts with 3D scene graphs, capturing the objects within a dynamic spatiotemporal graph in a 3D space \cite{cherian_251d_2022}. Closest to our setting, Khan et al. introduced situation hypergraphs, training a decoder to implicitly identify graph representations with actions and human-object relationships \cite{khan_learning_2023}. Finally, recent works explored hierarchical and relation-aware modeling. Qi et al.\cite{Qi} and Wang et al.\cite{Wang2023} emphasize relational propagation through a bottom-up hierarchical propagation over heterogeneous graphs and a dynamic relation-aware reasoning framework that combines appearance-motion and location-semantic object interactions respectively. Transformer-based graph models, proposed by Xiao et al.\cite{Xiao2022} can model long-range dependencies over dynamic graphs, while Tan et al.'s \cite{Tan2024} graph-based reasoning framework aims to capture both spatial and temporal object relations. These methods demostrate the value of structured graph reasoning, but often rely on deep propagation stages or transformer architectures that can be computationally heavy.

\paragraph{Relation to Prior Work}
Our approach departs from these works as it operates on explicit scene graphs and provides a lightweight and transparent reasoning process. Processing graphs compared to raw video frames enables us to reduce computational overhead, while using a hierarchical model allows us to capture video content in two granularities along with minimizing reliance on language. We introduce a human-rooted video-level graph: each frame has a local human root, and cross-frame links exist only via human nodes to a global root. A shallow heterogeneous-edge encoder (SGEM) and a hierarchical network then perform temporal reasoning over the sequence of human-centric embeddings. This human-anchored approach produces compact, interpretable representations for modeling human actions.

\section{Methodology}

\subsection{Problem Formulation}

This work follows a classification formulation for the VideoQA task as in \cite{agqa}. The objective is to predict the correct answer $a_i^* \in A$ over a fixed set of $K$ possible answers or $A = \{ a_1, a_2, ..., a_k\}$, for an input video $V_i \in \mathcal{V}$ representing a sequence of $N$ frames $V_i = [f_1, f_2, ..., f_N]$ and a question $q_i \in Q$. The goal becomes to learn a mapping function $\mathcal{F}: \mathcal{V} \times Q \rightarrow A$, that predicts the correct answer or $a_i^* = \mathcal{F}(V_i,q_i)$.

\subsection{Proposed Approach}

The mapping $\mathcal{F}$ is formed through a series of steps. 
\paragraph{Frame Scene Graph Generation}: In the core of GHR-VQA is an initial frame scene graph generation step for each individual frame of the input video. A scene graph $\mathcal{G}_i = (\mathcal{N}_i, \mathcal{E}_i)$ serves as a structured representation of a scene captured within the frame, where nodes $\mathcal{N}_i$ correspond to the entities or objects, and the edges $\mathcal{E}_i$ represent semantic relationships between these entities. Each node $n_j^i \in \mathcal{N}_i$ includes the bounding box coordinates of the entity and and the corresponding class label, while edges $e_{j,k}^i \in \mathcal{E}_i$ capture the type of semantic relationship between two objects $n_j^i$ and $n_k^i$. For example, given a frame showing a person holding a cup, the scene graph would consist of two nodes representing the "person" and the "cup" and an undirected edge labeled "holding" between them.

We start by processing each input frame $f_i$ individually via a Scene Graph Generation Module or $\mathcal{G}_i = \mathcal{S}_g(f_i)$. For $\mathcal{S}_g$ we use the off-the-shelf scene graph generator of ~\cite{SGDet}, which consists of a pre-trained detector backbone and is based on the scene generation module from MOTIFS~\cite{motifs}. $\mathcal{S}_g$ is trained on the Visual Genome dataset~\cite{vg} containing 150 object categories and 50 types of relationships/predicates. 

\begin{figure}[h!]
\begin{center}
   \includegraphics[width=0.8\linewidth]{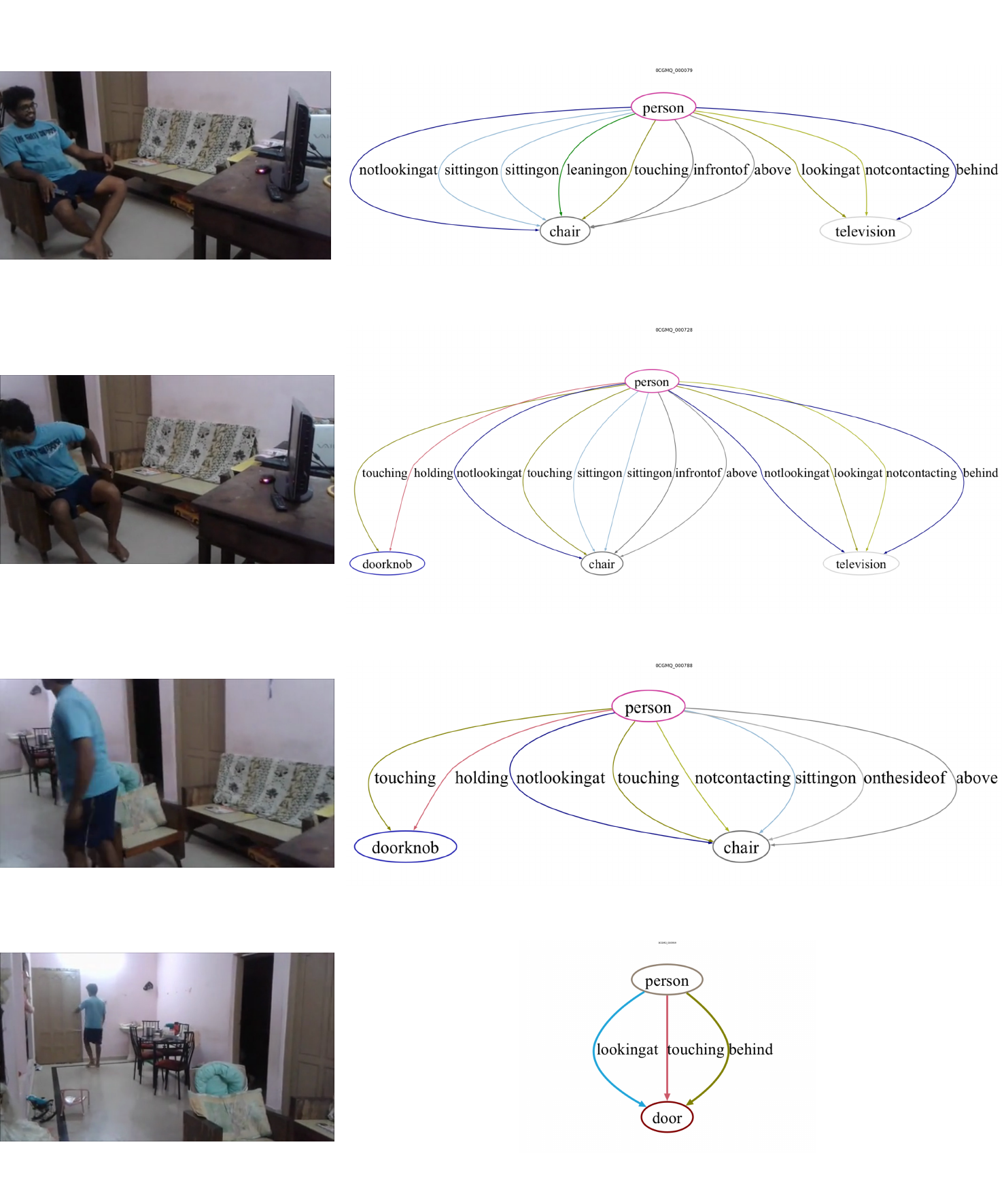}
\end{center}
   \caption{Example of 4 frames from the video sample OCGMQ with the corresponding annotated scene graphs.}
\label{fig:exampleframes}
\end{figure}
\paragraph{Video-level Graph Formulation}: To provide a bridge across frames, we designate one "human" node as the local root of each frame's graph. We construct a video-level human-rooted graph $\mathcal{G} = (\mathcal{N}, \mathcal{E})$ that links all frame-level graphs via their respective human nodes to a single video-level root node $n_\text{root}$. Formally $\mathcal{N} = \bigcup_{i=1}^N N_i \cup \{n_\text{root}\}$ and  $\mathcal{E} = \bigcup_{i=1}^N E_i \cup \{(n^i_\text{human}, n_\text{root}) ~|~ i = 1, ..., N\}$.

 This structure ensures that each human node is two steps away from any other node, enabling fast and compact cross-frame message passing in a human-centric scope. By restricting cross-frame connections to human nodes, the model is encouraged to base its reasoning on human-object relationships and their evolution through time.
\paragraph{Frame Scene Graph Encoding (SGEM)}: The generated video hypergraph is then encoded into a latent scene representation via a Scene Graph Encoding Module (SGEM) modeled as a 2-layer Heterogeneous Edge Graph Attention Network (HetEdgeGAT) to learn cross-entity semantic patterns. Our SGEM follows~\cite{monninger2023scene} where each node $n_j^i$ in graph $\mathcal{G}_i$ is updated as:

\begin{equation}
(n_j^i)' = \text{EdgeGAT}_r(n_j^i)= \Theta_{s,r} \cdot n_j^i + \Bigg\|_{h=1}^{H} \left( \sum_{k \in \mathcal{N}_r(n_j^i)} \alpha_{j,r,k}^h \left( \Theta_{v,r}^h \cdot n_k^i + \Theta_{e,r}^h \cdot e_{j,r,k} \right) \right)
\end{equation}

where $\Theta$ is used to denote learnable weight matrices for the transformation of features of the node to update self (s), neighboring nodes (u) and edge features (e) for graph $\mathcal{G}_i$. H corresponds to the number of attention heads and the $\|$ denotes the concatenation operator. Attention weights are obtained by: 

\begin{equation}
    \alpha_{j,r,k}^h = \text{softmax}_{r,k}\left( \text{LeakyReLU} \left( \right. \right. a_r^{h^T} \left[ \Theta_{v,r}^h \cdot n_j^i \; \| \; \Theta_{v,r}^h \cdot v_k^i \; \| \; \Theta_{e,r}^h \cdot e_{j,r,k} \right] \left. \left. \right) \right)
\end{equation}

with $\alpha$ corresponding to a learnable vector.

We use cascaded layers in order to aggregate information in the frame regarding all types of relationships. To get a comprehensive representation of the entire graph representation, we combine the individual embeddings of the local frame graph root nodes, the human nodes, into a single vector. This aggregation is achieved through summing the node embeddings for each human node  $n_j^i$ in the graph $\mathcal{G}_i$ are aggregated using the following function:  $\mathcal{G}_i = \mathcal{F} \left( \left\{ n_j^i | j \in \mathcal{N}_i \right\} \right)$ where $\mathcal{G}_i$ is the final graph embedding and $\mathcal{F}$ is a sum function $\mathcal{F} = \sum_{j \in N} n_j^i$. 

\paragraph{Question Encoder}
In addition to graph processing, we map the input question into a latent space by leveraging token-wise sentence embeddings extracted from the penultimate layer of a BERT model~\cite{devlin-etal-2019-bert}. Specifically, we utilize the [CLS] token embedding from the model’s output, for a holistic representation of the entire sentence.

\begin{table}[t]
\centering
\begin{adjustbox}{width=0.95\textwidth,center}
\begin{tabular}{lccccccccc}
\hline
\textbf{Method} & \textbf{obj-rel} & \textbf{rel-action} & \textbf{obj-action} & \textbf{superlative} & \textbf{sequencing} & \textbf{exists} & \textbf{duration} & \textbf{activity} & \multicolumn{1}{|c}{\textbf{Overall}} \\
\hline
\multicolumn{10}{c}{\textit{Results on Dataset Subset A (our subset)}} \\ \hline
PSAC \cite{PSAC} & 37.21 & 49.25 & 49.29 & 32.6 & 49.23 & 49.16 & 44.3 & 3.72 & \multicolumn{1}{|c}{39.68} \\
HME \cite{HME} & 36.7 & 49.1 & 49.35 & 33.32 & 49.28 & 49.26 & 46.33 & 5.43 & \multicolumn{1}{|c}{39.4} \\
HCRN \cite{HCRN} & 49.73 & 49.24 & 49.66 & 32.94 & 49.12 & 49.1 & 43.25 & 5.2 & \multicolumn{1}{|c}{41.78} \\
GHR-VQA (ours) & \textbf{49.8} & \underline{53.7} & \underline{55.1} & \textbf{40.9} & \underline{50.4} & \textbf{56.3} & 25.5 & \underline{7.4} & \multicolumn{1}{|c}{\underline{42.5}} \\
\hline
\multicolumn{10}{c}{\textit{Results on Dataset Subset B (full AGQA 2.0 dataset)}} \\ \hline
PSAC \cite{PSAC} & 37.84 & 49.95 & 50.00 & 33.20 & 49.78 & 49.94 & 45.21 & 4.14 & \multicolumn{1}{|c}{40.18} \\
HME \cite{HME} & 37.42 & 49.90 & 49.97 & 33.21 & 49.77 & 49.96 & \underline{47.03} & 5.43 & \multicolumn{1}{|c}{39.89} \\
HCRN \cite{HCRN} & 40.33 & 49.86 & 49.85 & 33.55 & 49.70 & 50.01 & 43.84 & 5.52 & \multicolumn{1}{|c}{42.11} \\
SHG-VQA \cite{khan_learning_2023} & \underline{46.42} & \textbf{60.67} & \textbf{64.63} & \underline{38.83} & \textbf{62.17} & \underline{56.06} & \textbf{48.15} & \textbf{10.12} & \multicolumn{1}{|c}{\textbf{49.20}} \\
\hline
\end{tabular}
\end{adjustbox}
\vspace{0.1cm}

\caption{Comparison of SOTA methods on different AGQA dataset subsets. Dataset Subset A corresponds to our experimental subset, while Dataset Subset B refers to the complete AGQA dataset.}
\label{tab:comparison_results}
\end{table}
\paragraph{Hierarchical Network}
We propose a hierarchical method to process graph embeddings at different levels of granularity to finally classify the answer to the question. Drawing inspiration from \cite{HCRN}, we adapt the model to integrate scene graphs, thus benefiting from the hierarchical and contextual processing of the CRN units. The core of the Hierarchical Network consists of multiple Conditional Relational Network (CRN) units arranged hierarchically. The CRN units at the lower level process data at the clip level, gathering information from multiple frames and handling more spatial information, whereas the CRN units at the higher level operate at video level, modeling longer temporal dependencies. The hierarchical design enables the model to consider information in different contexts. 

The top-level CRN layer outputs a video graph embedding, used to classify the answer. This video-graph embedding is aggregated with the question embeddings and is processed by an answer decoder that generates the final output.

Each CRN unit takes as input an array of $n$ objects $\mathcal{X} =(x_1,...,x_n),\ x_i \in \mathbb{R}^d$ and a conditioning feature $c$ as a global context. The objects correspond to the aggregated local root embedding from the $i$-th frame after processing by the SGEM. These embeddings encode the aggregated relational context centered on the human for that frame. The conditioning feature $c$ corresponds to the question encoding and is used by the CRN to compute a relation-based transformation $y_i$ : 
\begin{equation}
y_i = \mathcal{F}(x_i,c)    
\end{equation}

For a given relation order $k \in \mathbb{N} $ with $1 \leq k \leq n $ , the CRN samples $t \in \mathbb{N}$ subsets of size $k$ from $\mathcal{X}$. A sub-network processes each subset to extract intra-set relational features which are then conditioned on the question context $c$ by another sub-network.The outputs are finally aggregated to produce a single result vector $r^k$ which represents conditional relations for the entire video. 

\section{Experiments}

\subsection{Dataset}
We evaluate our method on the Action Genome Question Answering Dataset 2.0 \cite{agqa}. Action Genome Question Answering (AGQA) is a benchmark for compositional spatio-temporal reasoning. This benchmark contains 96.85M question-answer pairs and a balanced subset of 2.27M question-answer pairs. It targets compositional questions about actors, objects and their interactions, making it well-suited for evaluating human-centric reasoning.

The dataset comprises approximately 9.6k videos, each with a duration of 30 seconds, recorded at a frame rate of 30 frames per second (fps). This translates to an average of around 900 frames per video. However, a notable aspect of AGQA is the selective annotation process applied to these videos. The dataset's goal, extending Action Genome \cite{ji2020action}, is to decompose actions, so its focus is on annotating only the video segments where actions occur and objects are involved in it. Thus, 5 frames uniformly sampled across each action interval are annotated. For each annotated frame, there is information for the visible objects, their bounding boxes, their labels, and relationships between them. So, despite the large number of frames available per video, on average, only 35 frames per video are annotated. This approach underscores the dataset’s emphasis on specific, salient moments within the videos, rather than an exhaustive frame-by-frame annotation with redundant objects and relationships. Finally, another critical characteristic of AGQA is that it is single-actor centric, meaning that for each video, annotations focus on one person, and all object interactions are described relative to this actor.

\paragraph{Our experimental framework - Dataset Subset A}
To accommodate various computational capacities and enable faster while detailed analysis, we designed a distinct experimental framework. This framework was designed to maintain the original dataset's distribution through random sampling, ensuring an accurate representation of the AGQA dataset's diversity and complexity. Our training subset, mentioned in Table \ref{tab:comparison_results}  as Dataset Subset A, corresponds to 100K train question-answer pairs and 20K test question-answer pairs. We ensured no data leakage from the train to the test set by keeping different videos in each set. To validate the effectiveness of our experimental setup, we benchmarked it against the baseline models provided by the AGQA creators. Our evaluation confirms that these models achieve consistent performance across both the full dataset and our subset, serving as a grounding for subsequent analysis.

\subsection{Implementation Details}

In our experimental setup, we utilized PyTorch\cite{pytorch} as the main framework for all model training and development. For graph-related tasks, especially in Graph Neural Networks (GNNs), we employed the Deep Graph Library (DGL) \cite{DGL}, which is compatible with PyTorch and provides optimized graph data structures and operations. The SGEM and the Hierarchical network were trained end-to-end with cross-entropy over the fixed AGQA answer set. All our experiments were run using two machines, each equipped with four GPUs.

\section{Results \& Analysis}

\paragraph{Overall Performance} Considering the two AGQA evaluation settings reported in Table \ref{tab:comparison_results}, GHR-VQA achieves 42.5\% overall accuracy on our Dataset Subset A. With SHG-VQA achieving 49.2\% overall accuracy on the full AGQA Dataset (Dataset Subset B), our approach places second in overall score among the state-of-the-art methods listed across the two reported settings. Within Subset A, GHR-VQA achieves the highes overall accuracy among the baselines considered (PSAC, HME, HCRN).

\paragraph{Per-category insights} Within Subset A, GHR-VQA improves by 9.47\% on obj-rel category, 3.75\% on rel-action category, 5.1\% on obj-action, 7.35\% on superlative, 6.29\% on exists and 1.88\% on activity. Across splits, we observe the biggest improvement of 9.47\% absolute points on object-relation reasoning questions compared to the best baseline in that category and 3.38\% absolute points on the best-performing model on full AGQA (Dataset Subset B), SHG-VQA. 

We can infer that the large gains on relation-centric categories are consistent with out human-rooted video-level graph, where human-object relationships are explicitly models across frames. Our method's remarkable performance in ’exists’ and ’superlative’ categories suggests that it can accurately identify the occurence of concepts and objects, along with their order. However, we notice a drop in performance on the duration category, suggesting the need for stronger temporal cues beyond our lightweight graph pipeline.

\paragraph{Qualitative assessment} As seen in the annotated scene graphs of Figure  \ref{fig:exampleframes}, frame scene graphs can capture salient human-object interactions that the SGEM encodes into embeddings. More specifically, from each frame, the local human node becomes a root, whose embedding forms a human-centric representation. The model makes correct predictions in cases where the SGG captures salient human-object interactions as seen in the dataset-provided annotations of Figure \ref{fig:exampleframes}. On the contrary, failure cases arise when the SGG misses small or occluded objects or when relevant interactions span long temporal gaps.

\begin{table}[h]
\centering
\begin{tabular}{lc}
\toprule
\textbf{Model} & \textbf{Accuracy (\%)} \\
\midrule
GINE + MLP      & 32.8 \\
EdgeGAT  + MLP  & 34.6 \\
HetEdgeGAT + MLP & 33.9 \\
HetEdgeGAT + HCRN (GHR-VQA) & 42.5\\
\bottomrule
\end{tabular}
\vspace{0.1cm}
\caption{Accuracy comparison of different GNN architectures along with MLP or HCRN.}
\label{tab:gnn_accuracy}
\end{table}

\paragraph{Ablation Studies} Table \ref{tab:gnn_accuracy} provides a comparative analysis of the accuracy achieved by different GNN architectures integrated with either a simple MLP or our hierarchical model HCRN. With an MLP, GNN variants fall in a narrow range, 32.8-34.6\%, indicating that per-frame graph encoding alone yields limited gains regardless of architecture. Integrating
\emph{HetEdgeGAT} with HCRN improves accuracy to 42.5\%, showing that temporal reasoning over the sequence of
human-root embeddings is the primary reason of the improvement. SGEM introduces heterogeneous edge-based attention, effectively modeling diverse relationships between objects and actions and enabling richer semantic understanding, while the hierarchical network further enhances reasoning capabilities by capturing spatiotemporal dependencies across different levels. These findings validate the robustness and scalability of our proposed HetEdgeGAT + HCRN framework, confirming its ability to handle the complexity of visual reasoning tasks.

\section{Discussion \& Conclusion}

Our scene-graph guided hierarchical model for VideoQA achieves notable improvements in object-relation reasoning, outperforming existing methods in this category. Performance gains shown in relation-centric categories indicate that anchoring cross-frame reasoning on human nodes is effective for answering questions focused on human actions.

Representing videos as graphs explicitly anchored on humans moves towards modeling their behavior as the primary reasoning goal. Since nodes and edges correspond to actual entities and relations, the model's intermediate states can be inspected and explained. Operating on compact graphs decouples perception from reasoning, while each module can be upgraded independently. Also, graph inputs are lighter than dense video features, allowing their use on constrained hardware.

Human-centric interpretable video reasoning can applied to many diverse fields. Firstly, we can find use in assistive technologies from explaining actions and object use to visually impaired individuals, to assessing rehabilitation in clinical environments. In safety-critical settings, it can help with monitoring industrial or clinical environments. Another application could be in AR/VR step-wise guidance. In these settings, among others, graph-level explanations provide an actionable interface between perception and decision-making.

However, we need to mention several limitations of the study. The computational cost of scene graph generation can be heavy, especially for longer videos. Our performance is dependent on the quality of the generated scene graphs, as errors in object detection or relation labeling can propagate to the reasoning stage. When presented with multi-actor scenes, selecting a single local human root per frame can be ambiguous. Framing VideoQA as a classification over a fixed answer set simplifies evaluation but limits expressivity and real-world use. Finally, evaluation is limited to one benchmark, so broader validation is needed.

Future work should focus on addressing the outlined limitations. Firstly, we should aim to enhance temporal reasoning by integrating more advanced models for better sequencing or using temporal memory into the hierarchical network. About videos with multiple humans, future extensions could consider multi-root video graphs to jointly reason over several actors. Additionally, refining scene graph generation and improving scalability through techniques like graph sparsification could improve efficiency and performance. Finally, validation on additional VideoQA datasets, even with multi-actors videos could help assess domain generalization.

In conclusion, our scene-graph-guided hierarchical model, GHR-VQA, advances VideoQA by structuring videos as human-rooted video-level graphs and reasoning hierarchically over human-centric embeddings. This design moves beyond pixel-based approaches and incorporates interpretable relational reasoning through scene graphs. The results support that centering computation on human-object structure is an effective modeling prior for compositional video reasoning, providing a foundation for more nuanced and advanced systems.

\paragraph*{Acknowledgements} This research work was accomplished during D. Brilli’s diploma thesis, co-supervised by P. Maragos at NTUA, and D. Mallis and V. Pitsikalis at deeplab.ai. This co-supervision is part of Deeplab’s support of basic research to academic institutions, and was funded as far as D. Mallis, V. Pitsikalis, and any publication dissemination are concerned by deeplab.ai.

\bibliographystyle{bmvc2k_natbib}
\bibliography{egbib}

\end{document}